\documentclass{article} 
\usepackage{iclr2026_conference,times}

\usepackage{amsmath,amsfonts,bm}

\def\eqref#1{equation~\ref{#1}}

\def\1{\bm{1}}

\DeclareMathAlphabet{\mathsfit}{\encodingdefault}{\sfdefault}{m}{sl}
\SetMathAlphabet{\mathsfit}{bold}{\encodingdefault}{\sfdefault}{bx}{n}

\usepackage{hyperref}
\usepackage{url}
\usepackage[table]{xcolor}

\usepackage[most]{tcolorbox}
\usepackage{xcolor} 
\usepackage{enumitem} 
\usepackage{booktabs}
\usepackage{algorithm}
\usepackage{algpseudocode}
\usepackage{mathtools}

\usepackage{booktabs}
\usepackage[table]{xcolor}
\usepackage{tabularx}
\usepackage{array}

\usepackage{amsmath,amssymb,amsthm}

\theoremstyle{remark}

\newcommand{\sg}{\mathrm{sg}}

\title{AdvFD: Boosting Visual Generation via Adversarial Fréchet Distance Loss}

\newcommand{\equalmark}{\textsuperscript{*}}

\newcommand{\corrmark}{\textsuperscript{\textdagger}}

\author{
\begin{tabular}{c}
Mingju Gao$^{1,2}$\equalmark,
\quad
Jingkai Zhou$^{2}$\equalmark,
\quad
Kun Gai$^{2}$,
\quad
Changqian Yu$^{2}$\corrmark,
\quad
Hao Tang$^{1}$\corrmark
\\[0.6em]
{\normalfont
$^{1}$Peking University
\qquad
$^{2}$KlingAI Research
}
\\[0.7em]
{\normalfont
\textbf{Project Page: }\href{https://gasaiyu.github.io/AdvFD-page/}
{\texttt{gasaiyu.github.io/AdvFD-page}}
}
\end{tabular}
}

\iclrfinalcopy 
\begin{document}

\maketitle

\begingroup
\renewcommand{\thefootnote}{\fnsymbol{footnote}}
\footnotetext[1]{Equal contribution.\qquad \textsuperscript{\(\dagger\)}~Corresponding author.}
\footnotetext{Project Lead: Jingkai Zhou}
\endgroup

\begin{abstract}

Fréchet distance has recently emerged as an effective distribution-level objective for generator post-training, complementing the conventional sample-level diffusion and flow-matching losses. However, directly optimizing Fréchet objectives can cause \emph{Fréchet hacking}. The target metrics keep improving, but visual quality and Fréchet alignment in other feature spaces may stagnate or deteriorate. We attribute this failure to the static pretrained feature spaces used by existing Fréchet losses. These feature spaces provide incomplete and fixed views of the differences between real and generated distributions. To address this limitation, we propose \emph{Adversarial Fréchet Distance} (AdvFD), which complements the static representation targets in FD-Loss with a calibrated adversarially learned representation. AdvFD augments the original static Fréchet objective with a learnable representation that adversarially maximizes the Fréchet discrepancy between real and generated samples, while the generator minimizes the same discrepancy in the resulting adaptive feature space. To prevent the adversarial representation from trivially increasing the objective through feature amplification, we further introduce real-feature whitening, which normalizes its scale and covariance geometry and stabilizes the min--max optimization. Extensive experiments show that AdvFD consistently improves one-step generator post-training across both JiT and pMF backbones and across different model scales.

\end{abstract}

\section{Introduction}
\label{sec:intro}


Diffusion models~\citep{sohl2015deep,ho2020denoising,song2020score,rombach2022high} have become a dominant approach to high-fidelity visual generation. Flow-based methods~\citep{lipman2022flow,liu2022flow,albergo2025stochastic} provide a closely related view by transporting a simple prior toward the data distribution. Their objectives are defined over samples from the data distribution and the corresponding perturbation paths. In practice, the training process asks the model to predict the target noise, score, or velocity at sampled intermediate states. It does not explicitly minimize the discrepancy between their distribution and the real data distribution.

\begin{figure}[t]
\centering
\includegraphics[width=0.9\linewidth]{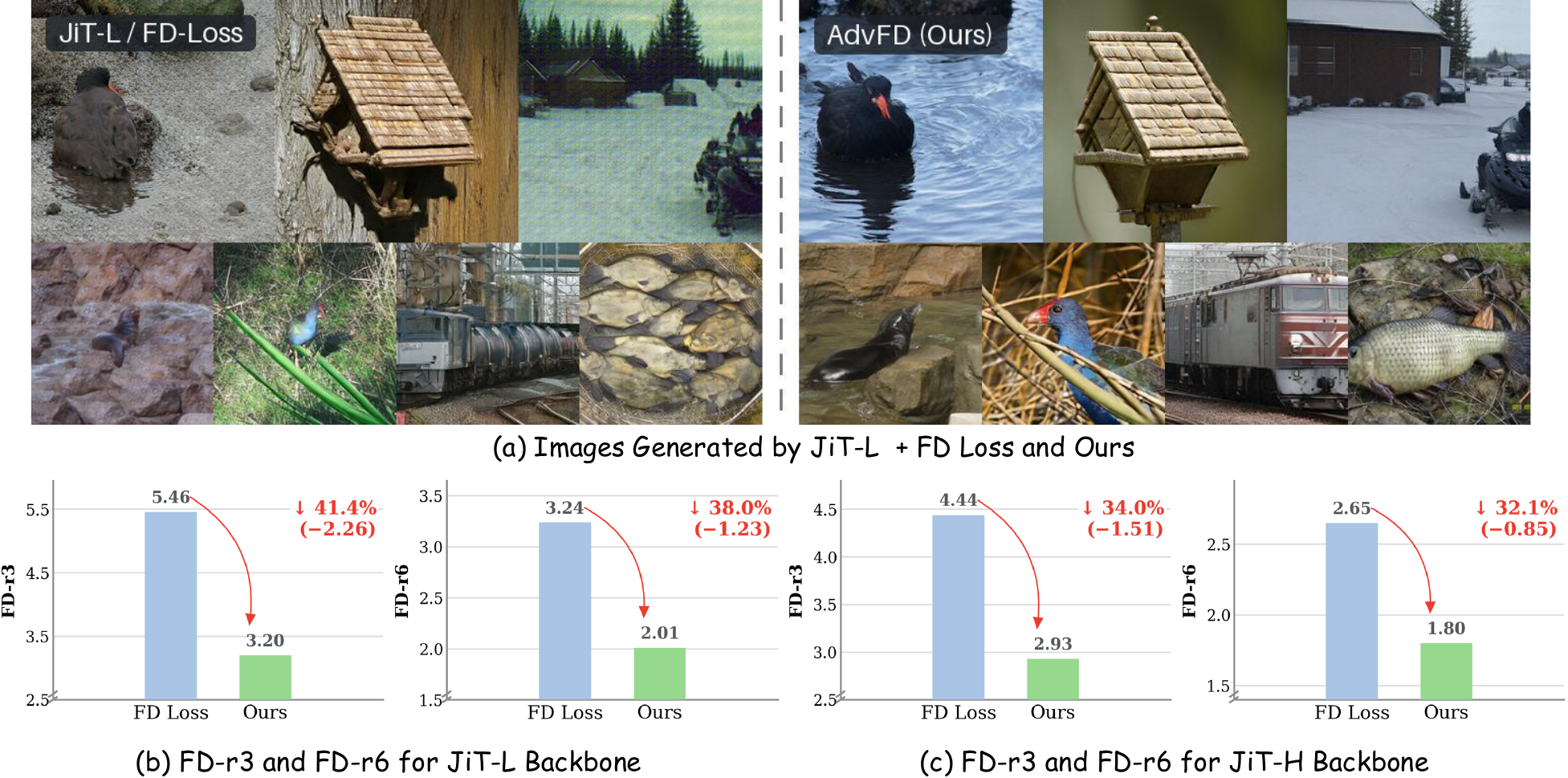}
\caption{
\textbf{AdvFD mitigates Fréchet hacking in one-step generation.}
\textbf{Top:} Compared with JiT-L trained using the static FD loss, AdvFD generates cleaner and more coherent images under the same 1-NFE sampling budget.
\textbf{Bottom:} AdvFD consistently reduces FD-r3 and FD-r6 across JiT-L and JiT-H, with relative improvements of 41.4\%/38.0\% and 34.0\%/32.1\%, respectively, showing that the gains persist as the generator scales up.
These results demonstrate improved perceptual quality and generalization across evaluation representations.
Lower is better for all metrics.}
\label{fig:teaser}
\end{figure}

To move beyond purely sample-level supervision, FD-loss~\citep{yang2026representation} introduces a distribution-matching objective for generator post-training. It minimizes the Fréchet distance between real and generated feature distributions extracted by pretrained encoders, such as Inception~\citep{szegedy2015going}, SigLIP~\citep{zhai2023sigmoid}, and MAE~\citep{he2022masked}, thereby aligning their first- and second-order statistics without requiring paired targets. Although FD-Loss performs well, its target metrics can keep improving as training continues, while visual quality and Fréchet scores from other encoders may remain unchanged or even worsen. As shown in Figure~\ref{fig:teaser} (a), post-training with FD-Loss introduces pronounced visual artifacts. Figure~\ref{fig:motivation} (b) quantifies the same issue: FD-r-Inception, the representation used for training, improves by 29.4\%, while FD-r-CLIP which is not used for training, worsens by 8.5\%.


\textbf{The Limits of Static Representation Targets.}
Why can the optimized metrics improve while visual quality and Fréchet scores from other encoders worsen? The issue arises because FD-Loss constrains the generator through predefined pretrained representations. Since each encoder captures only particular aspects of image semantics, structure, and texture, discrepancies outside its representation scope receive weak penalty. As optimization proceeds, the generator can become increasingly specialized to the selected feature spaces, continuing to reduce the target FD while preserving errors that remain visible to humans or other encoders. Figure~\ref{fig:motivation} (a) provides a direct example: a learned universal perturbation introduces visible high-frequency artifacts while reducing Inception FID. We refer to this behavior as \emph{Fréchet hacking}. Adding more frozen encoders may broaden the supervision, but the comparison spaces remain static. This motivates learning an adversarial representation that evolves with the generated distribution.



To address this issue, we propose \emph{Adversarial Fréchet Distance} (AdvFD), which complements the static encoder set in FD-Loss with a learnable representation. This representation is optimized to expose residual discrepancies between the real and generated distributions, while the generator continues to minimize a distribution-level Fréchet objective introduced by this representation. However, directly maximizing Fréchet distance with a trainable encoder can lead to trivial feature amplification and unstable optimization. We therefore introduce real-feature whitening to normalize the scale and covariance geometry of the adaptive feature space, yielding a dynamic yet stable representation target. AdvFD can be readily applied to one-step generators and consistently improves both training and cross-encoder Fréchet metrics.


We conduct extensive experiments on ImageNet class-conditional generation across multiple generators. As shown in Figure~\ref{fig:teaser}, AdvFD effectively reduces visual artifacts caused by optimizing static representation targets and consistently improves both the optimized Fréchet metrics and those computed with feature encoders not used during training. Our contributions are summarized as follows:
\begin{itemize}
\item We identify \emph{Fréchet hacking}, where optimizing static Fréchet objectives improves target metrics but can degrade visual quality and alignment under other representations.

\item We propose \emph{Adversarial Fréchet Distance} (AdvFD), which complements static FD targets with an adversarially learned representation that exposes residual real--generated discrepancies.

\item We introduce real-feature whitening to stabilize adversarial representation learning. AdvFD consistently improves generation quality across backbones and model scales, outperforming FD-loss and alternative remedies.
\end{itemize}





\section{Related Work}
\label{sec:related}

\subsection{Fréchet Distance in Visual Generation}

\textbf{Fréchet Distance as Evaluation.}
Fréchet-based distances have become standard tools for evaluating image generation~\citep{brock2018large,karras2019style,dhariwal2021diffusion,peebles2023scalable,ma2024sit,li2026back,zheng2025diffusion}.
Among them, the Fréchet Inception Distance (FID)~\citep{heusel2017gans} compares real and generated distributions through the first- and second-order statistics of features extracted from a pretrained Inception network~\citep{szegedy2015going}.
While FID is simple and widely correlated with visual quality, it is fundamentally tied to the Inception feature space and is affected by finite-sample estimation and the Gaussian approximation of feature distributions~\citep{binkowski2018demystifying,jayasumana2024rethinking}.
These limitations have motivated evaluation protocols that replace the underlying feature encoder or adopt alternative distributional discrepancies~\citep{jayasumana2024rethinking,yang2026representation,berthet2026mind}.


\textbf{Fréchet Distance as Training.} Prior work has incorporated feature-statistic discrepancies into adversarial objectives for GAN training. McGAN~\citep{mroueh2017mcgan} matches feature means and covariances through norm-based IPMs, Fisher GAN~\citep{mroueh2017fisher} normalizes a scalar critic by its second moment, MMD-GAN~\citep{li2017mmd} maximizes kernel MMD through a constrained encoder, and Fréchet GAN~\citep{doan2020image} directly computes Fréchet distance on discriminator features. Despite their different formulations, these methods are all developed for GANs and jointly learn the feature representation and generator from scratch. More recently, FD-Loss~\citep{yang2026representation} uses frozen pretrained encoders to provide a stable Fréchet training objective for post-training diffusion models. However, once these fixed representations become optimization targets, the generator can over-optimize their feature statistics while improving much more slowly under held-out representations, a phenomenon we term \emph{Fréchet hacking}. AdvFD extends Fréchet-based training to diffusion models by retaining the stable pretrained representations and introducing an adaptive branch that evolves with the generator. This allows the training objective to continuously expose residual distributional mismatches that are missed by fixed encoders, thereby mitigating Fréchet hacking and achieving stronger performance under both training and held-out representations.

\subsection{Mitigating Objective Hacking in Generative Training} 
Objective hacking arises when a generator over-optimizes an imperfect proxy, improving the measured objective by exploiting unpenalized artifacts rather than improving the intended visual quality~\citep{pan2022effects,gao2023scaling}.
Existing mitigation strategies can be broadly grouped into pretraining-stage and post-training approaches. 
During pretraining-stage distribution matching, prior works alleviate the limitations of incomplete supervision through adaptive feedback, including patch-level critics, pretrained-feature discriminators, teacher-based distribution matching, and adversarial diffusion distillation~\citep{isola2017image,sauer2021projected,kumari2022ensembling,yin2024one,yin2024improved,sauer2024adversarial,sauer2024fast,lu2025adversarial}.
In post-training alignment, models directly optimize learned aesthetic, preference, or alignment rewards, which can lead to reward over-optimization; recent methods mitigate this by calibrating reward confidence, constraining reward optimization, or adversarially updating reward models against generated samples~\citep{kim2024confidence,zhai2025mira,mao2026image}. 
While these methods reduce proxy exploitation through external critics, rewards, or distillation signals, AdvFD targets Fréchet hacking in pretrained feature-space distribution matching itself by calibrating the representation space.


\section{Fréchet Hacking under Static Representation Matching}
\label{sec:motivation}

\begin{figure}[t]
\centering
\includegraphics[width=\linewidth]{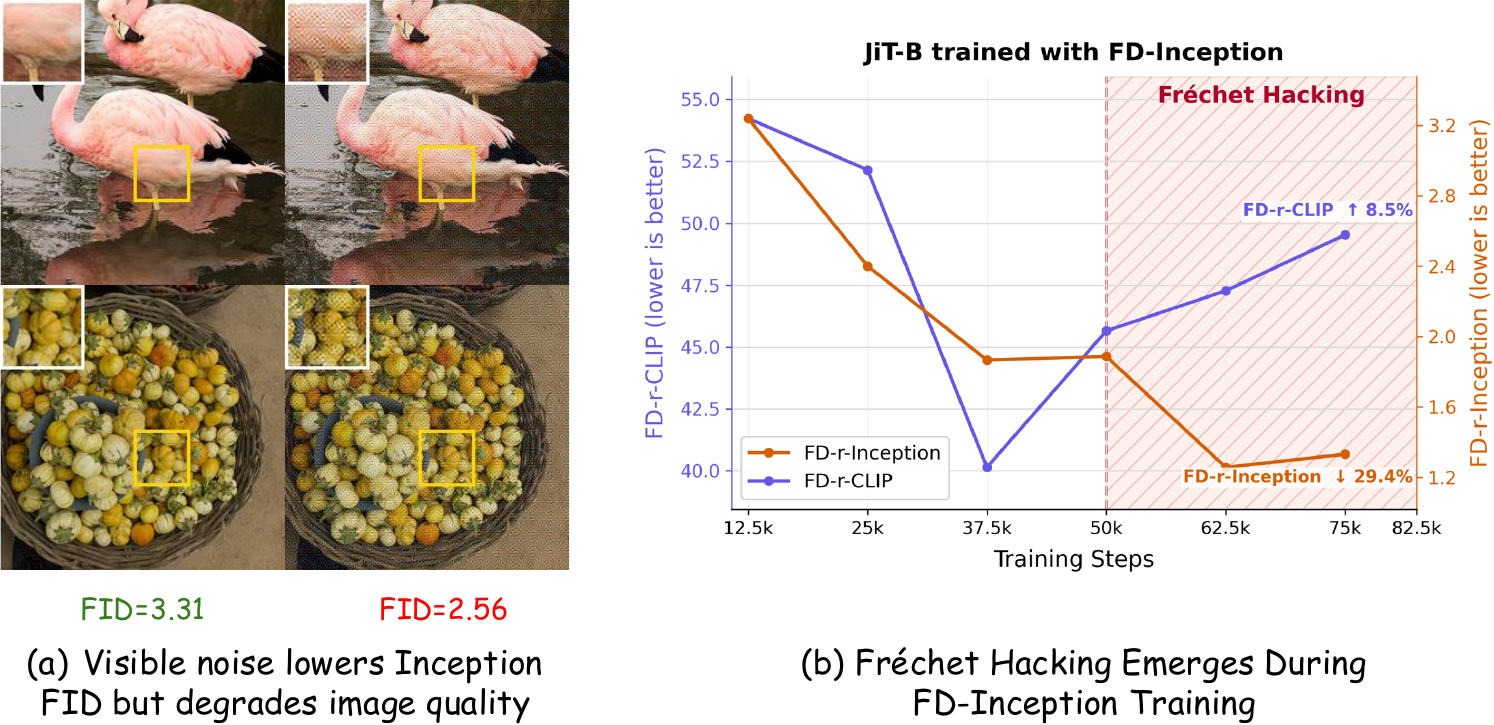}
\caption{
\textbf{Fréchet hacking under a static Inception representation.}
\textbf{Left:} A universal learned perturbation optimized solely for Inception FID introduces visible high-frequency artifacts while reducing FID from 3.31 to 2.56, directly revealing an exploitable blind direction of the static representation.
\textbf{Right:} The same failure mode emerges during JiT-B post-training: from 50k-75k steps, FD-r-Inception decreases by 29.4\%, whereas FD-r-CLIP increases by 8.5\%.
Together, these results show that optimizing a static Fréchet objective can improve the target metric while degrading visual quality and distributional alignment in a feature space not used for training.
}

\label{fig:motivation}
\end{figure}

Feature-space distribution matching provides a direct training objective for visual generation by comparing real and generated distributions without paired reconstruction targets. FD-Loss~\citep{yang2026representation} implements this idea by minimizing the Fréchet distance between their feature distributions, thereby aligning first- and second-order statistics without teacher distillation.
Results show that directly optimizing such representation-space distribution distances can effectively improve generator post-training. 
However, this success rests on an implicit assumption:
\begin{tcolorbox}[
  enhanced,
  colback={rgb,255:red,247;green,247;blue,247},
  colframe={rgb,255:red,0;green,86;blue,80},
  boxrule=1.0pt,
  arc=5pt,
  left=14pt,
  right=14pt,
  top=8pt,
  bottom=8pt,
  boxsep=0pt,
  width=\linewidth,
  before skip=8pt,
  after skip=8pt,
  drop fuzzy shadow
]
\noindent\textbullet\quad
\textbf{Current Assumption:}
\textit{Aligning generated and real samples in a static pretrained feature space could improve visual quality.}
\end{tcolorbox}
However, this assumption does not always hold. The generator is supervised only by the static feature spaces, so it mainly corrects the discrepancies that these representations can detect. Differences they fail to capture may remain weakly constrained, allowing important visual structures and semantic details to degrade during training. As a result, the optimized FD may continue to decrease even when visual quality and alignment in other feature spaces, such as CLIP, stagnate or worsen.

We illustrate the limitations of using the static Inception representation in Figure~\ref{fig:motivation}. \textbf{Left:} We freeze a pretrained pMF-B generator and learn a universal perturbation solely by minimizing its Inception FID. Although the learned noise introduces clearly visible artifacts, it reduces FID from 3.31 to 2.56, demonstrating that the static representation contains directions along which visual quality can deteriorate while the target metric improves. \textbf{Right:} The same limitation appears during FD-Inception post-training: from 50k to 75k steps, FD-r-Inception decreases by 29.4\%, whereas FD-r-CLIP increases by 8.5\%. These results show that a static feature space can overlook perceptually relevant discrepancies and provide a misleading optimization signal.

\textbf{From Static to adversarial representations.}
These observations motivate extending the representation scope of Fréchet training beyond a fixed encoder set. Although adding more frozen encoders can broaden feature coverage, it also increases computational and memory costs while remaining limited to predefined representations. More importantly, a static ensemble may still overlook discrepancies that emerge as the generator adapts to the training objectives. We therefore introduce a learnable representation that is updated using the current real and generated distributions. The representation enlarges their Fréchet discrepancy to expose residual mismatches, while the generator minimizes the same discrepancy in the resulting feature space. As the generated distribution evolves, the representation adapts accordingly, providing dynamic and complementary supervision beyond the fixed encoders.

\section{Method}
\label{sec:method}

\begin{figure}[t]
\centering
\includegraphics[width=0.95\linewidth]{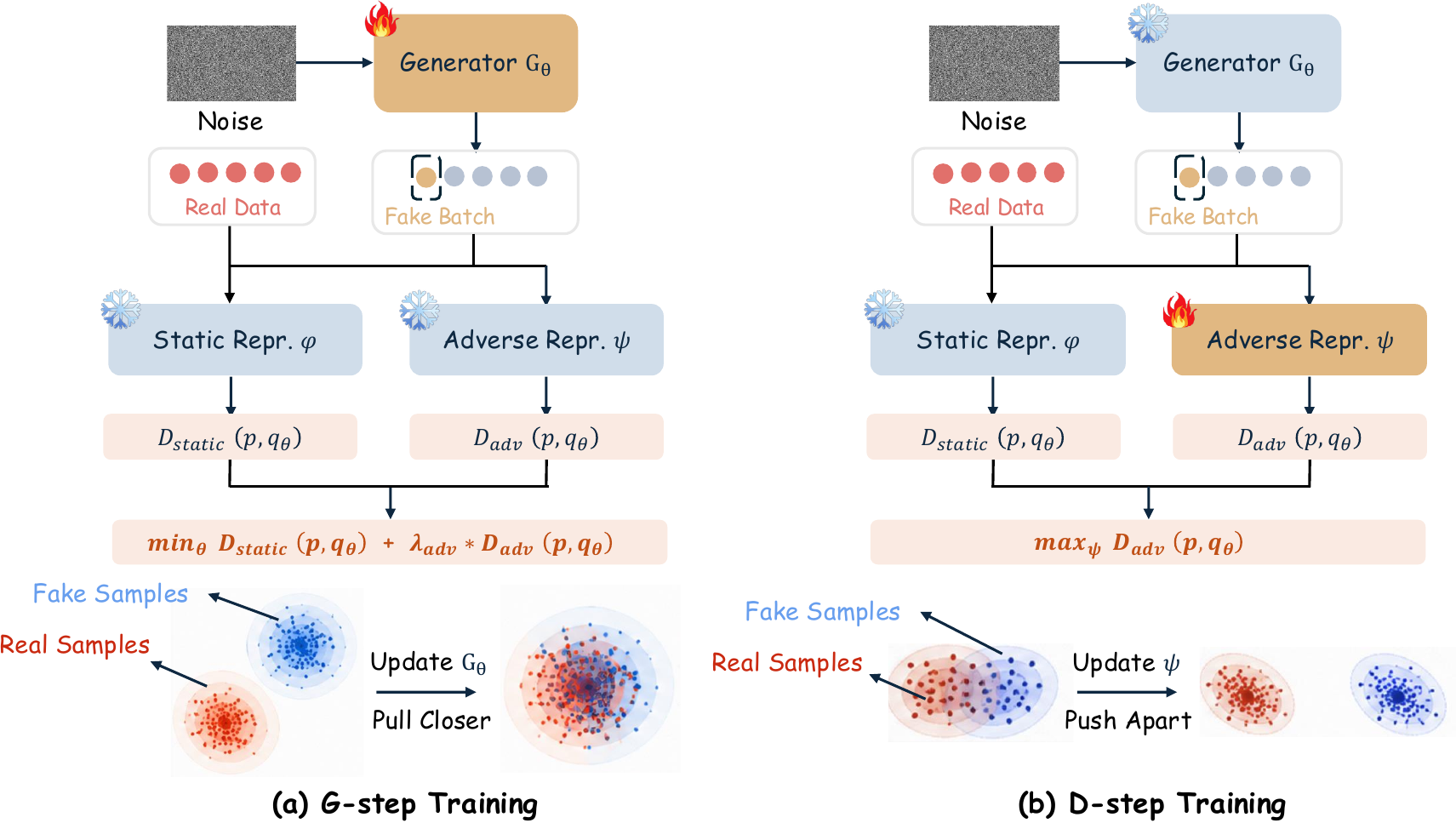}
\caption{
\textbf{Overview of our adaptive training.}
(a) In the \textbf{G-step}, the generator is updated to minimize the static Fréchet objective together with the adaptive Fréchet discrepancy, while both the static and adversarial representations are frozen.
(b) In the \textbf{D-step}, the generator is frozen, and the adversarial representation is updated to maximize the Fréchet discrepancy between real and generated distributions.
The two steps alternately improve the generator and adapt the representation to discrepancies missed by the fixed encoder.
}
\label{fig:method}
\end{figure}

\subsection{Adaptive Fréchet Loss}
\label{sec:AdvFD}

Let $p$ denote the real data distribution, $q_{\theta}$ the distribution induced by the generator $G_{\theta}$, and $\mathcal{F}$ a set of frozen visual representations, including SigLIP~\citep{zhai2023sigmoid}, MAE~\citep{he2022masked}, and Inception~\citep{szegedy2015going}. Following FD-Loss, we define the static Fréchet objective as
\begin{equation}
\label{eq:fd-static}
D_{\mathrm{static}}\left(p,q_{\theta}\right)=\sum_{\phi\in\mathcal{F}}\lambda_{\phi}D_{\mathrm{FD}}^{\phi}\left(p,q_{\theta}\right),
\end{equation}
where $\lambda_{\phi}$ controls the contribution of each representation. These frozen feature spaces provide stable distribution-level supervision but may overlook residual discrepancies as the generated distribution evolves. We therefore introduce a trainable representation $\psi_{\omega}$, initialized from a pretrained visual encoder, to adapt the comparison space to the current generator. 

At iteration $t$, the current representation $\psi_{\omega_t}$ defines an adaptive Fréchet term that complements the static objective. Its calibrated form is introduced below. With $\omega_t$ fixed, the generator minimizes
\begin{equation}
\label{eq:adafd-minmax}
\mathcal{L}_{t}\left(\theta;\omega_t\right)
=
D_{\mathrm{static}}\left(p,q_{\theta}\right)
+
\lambda_{\mathrm{adv}}
D_{\mathrm{adv}}\left(p,q_{\theta};\omega_t\right).
\end{equation}
AdvFD alternates between two updates. The generator reduces the discrepancy measured by the current representation, while the adversarial representation increases $D_{\mathrm{adv}}$ to reveal remaining differences between the real and generated distributions. This forms a GAN-like game in feature space. The complete training procedure is summarized in Algorithm~\ref{alg:advfd} and Figure~\ref{fig:method}.

\paragraph{G-step: updating the generator.}
As shown in Figure~\ref{fig:method} (a), the frozen representations in $\mathcal{F}$ and the adversarial representation $\psi_{\omega_t}$ are fixed during the G-step. Gradients are propagated only through the generated samples, and the generator is updated as
\begin{equation}
\label{eq:g-step-update}
\theta_{t+1}
=
\theta_t
-
\eta_{\mathrm{G}}
\left.
\nabla_{\theta}
\left[
D_{\mathrm{static}}\left(p,q_{\theta}\right)
+
\lambda_{\mathrm{adv}}
D_{\mathrm{adv}}\left(p,q_{\theta};\omega_t\right)
\right]
\right|_{\theta=\theta_t},
\end{equation}
where $\eta_{\mathrm{G}}$ is the generator learning rate. The static term maintains alignment in the pretrained feature spaces, while the adaptive term reduces the residual discrepancy exposed by the current learned representation.

\paragraph{D-step: updating the adversarial representation.}
As shown in Figure~\ref{fig:method} (b), the updated generator is frozen during the D-step, and generated samples are detached from the computational graph. We first compute
\begin{equation}
\label{eq:d-step-gradient}
g_t=\nabla_\omega D_{\mathrm{adv}}\big(X_r,\,\sg\!\big(G_{\theta_{t+1}}(Z)\big);\omega\big)\Big|_{\omega=\omega_t}
\end{equation}
The representation increment can be equivalently characterized as the solution to the locally constrained response problem
\begin{equation}
\label{eq:d-step-local-response}
\Delta_t
=
\underset{\|\Delta\|_2\leq\eta_{\mathrm{D}}\tau}{\arg\max}
\left\{
\langle g_t,\Delta\rangle
-
\frac{1}{2\eta_{\mathrm{D}}}
\|\Delta\|_2^2
\right\}.
\end{equation}
This objective maximizes the first-order increase in the adaptive discrepancy while penalizing large departures from the current representation. Its closed-form solution motivates the gradient clipping used in our D-step:
\begin{equation}
\label{eq:d-step-update}
\omega_{t+1}
=
\omega_t
+
\eta_{\mathrm{D}}
\operatorname{clip}_{\tau}\left(g_t\right),
\qquad
\operatorname{clip}_{\tau}(g)
=
g\min\left(1,\frac{\tau}{\lVert g\rVert_2}\right).
\end{equation}
Here, $\eta_{\mathrm{D}}$ is the representation learning rate and $\tau$ is the gradient-norm clipping threshold. We implement this bounded response with AdamW~\citep{loshchilov2017decoupled} under the same clipping threshold.
Each D-step selects a bounded local response that increases the real--generated discrepancy around the current representation. The G-step and D-step are applied alternately throughout post-training.

\paragraph{Preventing Feature-Scale Explosion.}
Directly increasing raw FD in a trainable feature space admits a trivial scale direction. In particular,
\begin{equation}
\label{eq:fd-scale-degeneracy}
D_{\mathrm{FD}}^{c\psi_{\omega}}
\left(
p,q_{\theta}
\right)
=
c^{2}
D_{\mathrm{FD}}^{\psi_{\omega}}
\left(
p,q_{\theta}
\right),
\qquad c>0.
\end{equation}
Thus, the adversarial representation could increase raw FD simply by enlarging its feature norm, without exposing additional differences between the real and generated distributions.

We remove this coordinate degeneracy through real-feature whitening. Let $\mu_{p}^{\psi}$ and $\Sigma_{p}^{\psi}$ denote the mean and covariance of the real features under the current representation $\psi_{\omega}$. We define
\begin{equation}
\label{eq:real-whiten}
\bar{\psi}_{\omega}(x)
=
\left(
\psi_{\omega}(x)-\mu_{p}^{\psi}
\right)
\left(
\Sigma_{p}^{\psi}+\epsilon I
\right)^{-1/2},
\end{equation}
where $\epsilon>0$ regularizes low-variance directions. In the full-rank population setting with $\epsilon=0$, affine-equivalent representations are mapped to whitened representations that differ only by an orthogonal transformation, under which FD is invariant. Appendix~\ref{app:whitening-derivation} provides the detailed derivation.

The adaptive discrepancy used in the G-step and D-step is then defined as
\begin{equation}
\label{eq:fd-adaptive}
D_{\mathrm{adv}}\left(p,q_{\theta};\omega\right)
=
D_{\mathrm{FD}}^{\bar{\psi}_{\omega}}
\left(p,q_{\theta}\right).
\end{equation}
It measures generated-feature deviations relative to the scale and covariance geometry of the real distribution. Under exact whitening, a common rescaling of $\psi_{\omega}$ is canceled by the corresponding change in the real-feature statistics. Therefore, $D_{\mathrm{adv}}$ cannot be increased solely through global feature rescaling. Appendix~\ref{app:whitening-derivation} analyzes the regularized form used in practice.

\begin{algorithm}[t]
\caption{AdvFD Training Procedure}
\label{alg:advfd}
\begin{algorithmic}[1]
\Require Pretrained generator $G_{\theta}$, frozen representations $\mathcal{F}$, pretrained adversarial representation $\psi_{\omega}$, real distribution $p$, adaptive weight $\lambda_{\mathrm{adv}}$, learning rates $\eta_{\mathrm{G}}$ and $\eta_{\mathrm{D}}$, clipping threshold $\tau$, and training iterations $T$
\For{$t=0,\ldots,T-1$}
    \State Sample a real batch $X_r\sim p$ and noise $Z\sim p_Z$
    \State Generate $X_g\gets G_{\theta_t}(Z)$
    \Statex
    \State \textbf{G-step: update the generator}
    \State Freeze the frozen representations $\mathcal{F}$ and adversarial representation $\psi_{\omega_t}$
    \State Compute $D_{\mathrm{static}}\gets\sum_{\phi\in\mathcal{F}}\lambda_{\phi}D_{\mathrm{FD}}^{\phi}(X_r,X_g)$
    \State Compute $D_{\mathrm{adv}}\gets D_{\mathrm{FD}}^{\bar{\psi}_{\omega_t}}(X_r,X_g)$ using real-feature whitening
    \State $\theta_{t+1}\gets\theta_t-\eta_{\mathrm{G}}\nabla_{\theta}\left[D_{\mathrm{static}}+\lambda_{\mathrm{adv}}D_{\mathrm{adv}}\right]$
    \Statex
    \State \textbf{D-step: update the adversarial representation}
    \State Freeze $G_{\theta_{t+1}}$ and detach the generated samples
    \State $X_g^{\mathrm{det}}\gets\operatorname{sg}\!\left[G_{\theta_{t+1}}(Z)\right]$
    \State Recompute $D_{\mathrm{adv}}\gets D_{\mathrm{FD}}^{\bar{\psi}_{\omega_t}}(X_r,X_g^{\mathrm{det}})$
    \State $g_t\gets\nabla_{\omega}D_{\mathrm{adv}}$
    \State $\omega_{t+1}\gets\omega_t+\eta_{\mathrm{D}}\,g_t\min\left(1,\frac{\tau}{\lVert g_t\rVert_2}\right)$
\EndFor
\State \Return post-trained generator $G_{\theta_T}$
\end{algorithmic}
\end{algorithm}

Algorithm~\ref{alg:advfd} summarizes the alternating training procedure of AdvFD. During the G-step, both the static and adversarial representations are frozen, while the generator is updated to minimize the corresponding Fréchet discrepancies. This step pulls the generated distribution closer to the real distribution in the current feature spaces, as illustrated in Figure~\ref{fig:method} (a). During the D-step, the generator is frozen, and the adversarial representation is updated to maximize the Fréchet discrepancy, effectively pushing the real and generated feature distributions apart to expose residual differences that are not captured by the static representations, as shown in Figure~\ref{fig:method} (b). By alternating between pulling the distributions together and adaptively pushing them apart, AdvFD provides the generator with a continuously evolving distribution-level training signal.

\subsection{Relation to GANs}
\label{sec:rel_with_gans}

Following the variational view of adversarial training, both GANs and AdvFD optimize a discrepancy through a nested inner--outer problem:
\begin{equation}
\label{eq:general-adversarial-objective}
\min_{\theta}
\sup_{u\in\mathcal{U}}
\Delta_{u}
\left(
p,q_{\theta}
\right).
\end{equation}
For a static generated distribution $q_{\theta}$, the inner optimization adapts $u$ to expose the remaining mismatch between $p$ and $q_{\theta}$. The outer optimization then updates the generator to reduce the discrepancy identified by the inner player. As $q_{\theta}$ evolves, the inner player is updated accordingly, allowing both methods to provide an adaptive distribution-matching signal.

The key distinction is what the inner optimization learns. In WGAN, the ground cost $c_X(x,y)$ in the original input space is fixed before training. The critic $f(x)$ then assigns a scalar score to each sample and is optimized to give real and generated samples different scores, yielding the Kantorovich dual formulation:
\begin{equation}
\label{eq:wgan-transport-view}
W_{1}
\left(
p,q_{\theta}
\right)
=
\sup_{\|f\|_{\mathrm{Lip}}\leq 1}
\left[
\mathbb{E}_{x\sim p}[f(x)]
-
\mathbb{E}_{x\sim q_{\theta}}[f(x)]
\right].
\end{equation}
The critic changes these scalar scores, but it does not change the distance between samples or the transport cost used to define the problem. AdvFD uses a different mechanism. It does not assign a scalar score to each sample; instead, the representation network $\psi_{\omega}$ maps each sample to a feature vector. For real and generated samples, AdvFD computes the mean and covariance of these feature vectors and compares the corresponding moment-matched Gaussian distributions using a fixed closed-form $W_{2}^{2}$ expression:
\begin{equation}
\label{eq:adafd-adversarial-view}
D^{\psi_\omega}_{\mathrm{FD}}(p,q_\theta)=W_2^2\Big(\mathcal N\big(\mu^{\psi}_{p},\Sigma^{\psi}_{p}\big),\,\mathcal N\big(\mu^{\psi}_{q},\Sigma^{\psi}_{q}\big)\Big)
\end{equation}
where $\mu_{p}^{\psi}$ and $\Sigma_{p}^{\psi}$ are the mean and covariance of $\psi_{\omega}(x)$ for real samples, and $\mu_{q}^{\psi}$ and $\Sigma_{q}^{\psi}$ are the corresponding statistics for generated samples. Learning $\psi_{\omega}$ therefore changes which differences between real and generated samples are reflected in these feature statistics, while the formula used to compare the statistics remains unchanged. In other words, \textbf{WGAN uses a fixed geometry with a learned transport potential, whereas AdvFD uses a learned geometry with a fixed Gaussian transport functional.}

\section{Experiment}
\label{sec:exp}

\subsection{Settings}

\textbf{Datasets and Training Protocol.}
We evaluate AdvFD on class-conditional image generation using ImageNet-1K~\citep{russakovsky2015imagenet} at a resolution of $256\times256$. Specifically, we consider JiT~\citep{li2026back} and pixel MeanFlow (pMF)~\citep{lu2026one} as generators, including both B-, L- and H-scale variants.

Following the post-training protocol of FD-loss~\citep{yang2026representation}, all methods are initialized from the same pretrained checkpoints and optimized on the ImageNet-1K training set. For each backbone, we compare the original generator, the corresponding FD-loss baseline, and AdvFD under the same training data, global batch size, optimization budget, and sampling configuration. Unless otherwise specified, AdvFD uses the same static Fréchet representations as its FD-loss baseline and introduces only the additional adversarial representation. During evaluation, all one-step models generate 50,000 samples using identical sampling settings to ensure a fair comparison.

\textbf{Evaluation Metrics.}
We report FID, FD-r6, and FD-r3. Following the evaluation protocol of FD-loss~\citep{yang2026representation}, FID measures distributional discrepancy in the Inception feature space, while FD-r6 averages normalized Fréchet distance ratios across six visual representations: Inception~\citep{szegedy2015going}, ConvNeXt~\citep{liu2022convnet}, DINOv2~\citep{oquab2023dinov2}, MAE~\citep{he2022masked}, SigLIP~\citep{zhai2023sigmoid}, and CLIP~\citep{radford2021learning}. Since the SIM (\textbf{S}igLIP+\textbf{I}nception+\textbf{M}AE) objective used for training includes SigLIP, MAE, and Inception, we additionally report FD-r3 over the remaining three encoders, namely ConvNeXt, DINOv2, and CLIP. FD-r3 therefore evaluates whether improvements in the optimized feature spaces transfer to representations not used during training. Lower values indicate better distribution matching for all metrics.

\hypersetup{hidelinks}

\definecolor{fdrow}{RGB}{242,239,232}
\definecolor{oursrow}{RGB}{232,243,250}

\newcommand{\methodcite}[1]{}


\newcommand{\improve}[1]{%
  \,\textcolor{red}{\scriptsize($\downarrow$#1\%)}%
}


\subsection{Main Results on C2I}

\begin{figure}[t]
\centering
\includegraphics[width=\linewidth]{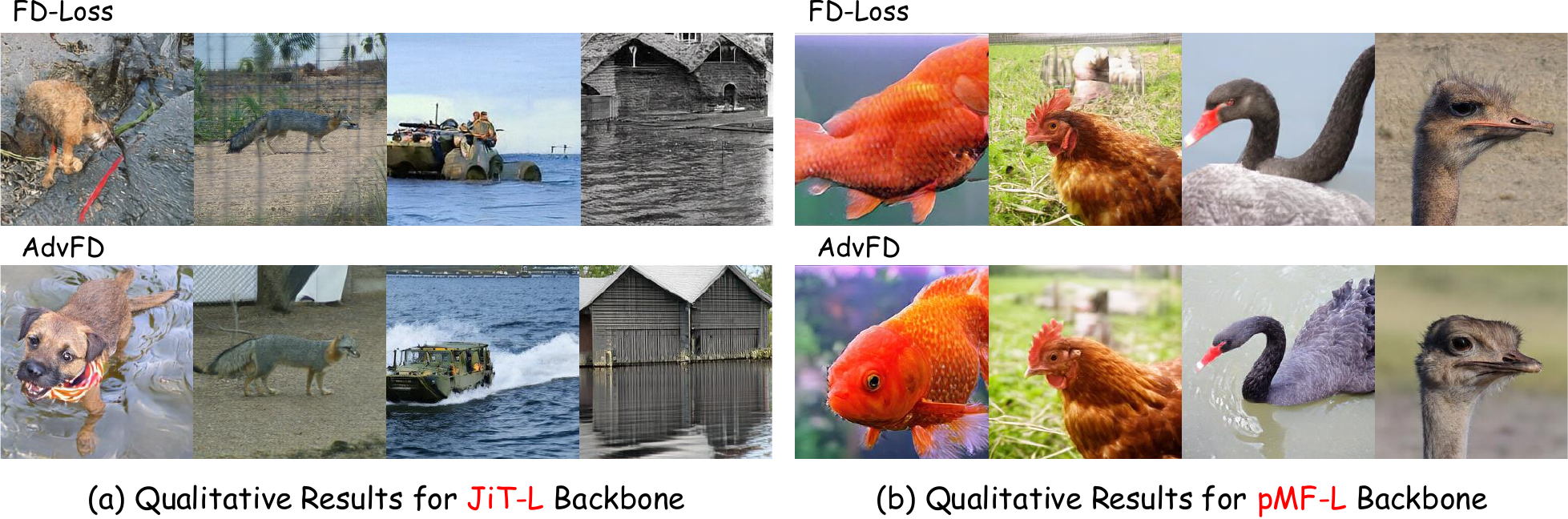}
\caption{
\textbf{Qualitative comparison between FD-Loss and AdvFD on ImageNet $256{\times}256$.}
We show representative samples generated by JiT-L (left) and pMF-L (right). Compared with FD-Loss, AdvFD produces cleaner textures, more coherent object structures, and fewer visible artifacts across both backbones.
}
\label{fig:quali}
\end{figure}

\textbf{Quantitative Results.} Table~\ref{tab:imagenet256_main} compares AdvFD with representative  pixel-space generative models on ImageNet $256\times256$. Across all JiT and pMF scales (B/L/H), AdvFD consistently outperforms the corresponding FD-loss baseline while preserving one-step generation. On JiT-B, AdvFD reduces FID from 1.00 to 0.79, FD-r6 from 5.53 to 3.92, and the held-out FD-r3 from 8.45 to 6.03, corresponding to relative improvements of 21.0\%, 29.1\%, and 28.6\%, respectively. The gains are more pronounced on JiT-L, where AdvFD achieves an FID of 0.73, an FD-r6 of 2.01, and an FD-r3 of 3.20, improving upon FD-loss by 5.2\%, 38.0\%, and 41.4\%, respectively. These gains persist at the H scale: on JiT-H, AdvFD improves FID, FD-r6, and FD-r3 from 0.75/2.65/4.44 to 0.72/1.80/2.93, corresponding to relative improvements of 4.0\%, 32.1\%, and 34.0\%. AdvFD also consistently improves all three metrics across pMF-B/L/H; on pMF-H, it further improves 0.77/1.89/2.69 to 0.74/1.74/2.50. Notably, the consistent reductions in FD-r3 across architectures and scales indicate that the improvements generalize to representations excluded from training, rather than being confined to the optimized SIM feature spaces. Overall, these results demonstrate that AdvFD generalizes across both generator architectures and model scales.

\textbf{Qualitative Results.}
Figure~\ref{fig:quali} provides qualitative comparisons on representative JiT-L and pMF-L backbones. Across both architectures, AdvFD produces more coherent object structures, cleaner textures, and fewer local artifacts than FD-Loss. These qualitative improvements are consistent with the quantitative gains, demonstrating that AdvFD generalizes across different generator architectures.

\begin{table*}[!t]
\centering

\caption{
\textbf{Class-conditional generation on ImageNet $256{\times}256$.}
We compare AdvFD with discrete-space, latent-space, and pixel-space
generative models.
FD-loss and AdvFD use the SIM representations, including SigLIP,
Inception, and MAE.
FID and FD-r6 follow the FD-loss evaluation protocol, while FD-r3
excludes the SIM representations used during training to evaluate
generalization beyond the optimized representation set.
Lower is better for all metrics. 
$^\dagger$For methods using interval CFG, we report the full-CFG NFE upper bound following FD-Loss.
}
\label{tab:imagenet256_main}

\setlength{\tabcolsep}{4.5pt}
\renewcommand{\arraystretch}{0.90}
\setlength{\aboverulesep}{0.35ex}
\setlength{\belowrulesep}{0.35ex}

\scriptsize

\resizebox{0.85\textwidth}{!}{%
\begin{tabular}{@{}lcccccc@{}}
\toprule
Method
& NFE
& Space
& \#Params
& FID $\downarrow$
& FD-r6 $\downarrow$
& FD-r3 $\downarrow$ \\
\midrule

\multicolumn{7}{@{}l}{\textit{reference (real images)}} \\

\textcolor{gray}{50k validation images}
& \textcolor{gray}{N/A}
& \textcolor{gray}{N/A}
& \textcolor{gray}{N/A}
& \textcolor{gray}{1.68}
& \textcolor{gray}{1.00}
& \textcolor{gray}{1.00} \\

\midrule

\multicolumn{7}{@{}l}{\textit{discrete-space models}} \\

VAR-d30\methodcite{var}
& $10{\times}2$
& discrete
& 2B
& 1.97
& 6.70
& 6.77 \\

BAR-L\methodcite{bar}
& $256{\times}2{\times}4$
& discrete
& 1.1B
& \textbf{1.01}
& \textbf{3.57}
& \textbf{3.35} \\

\midrule

\multicolumn{7}{@{}l}{
  \textit{latent-space models, multi-step}
} \\

\multicolumn{7}{@{}l}{
  \quad\textit{without semantic distillation}
} \\

SiT-XL/2\methodcite{sit}
& $250{\times}2$
& latent
& 675M
& 2.12
& 8.44
& 9.20 \\

MAR-L\methodcite{mar}
& $256{\times}2{\times}100$
& latent
& 478M
& 1.80
& 6.68
& 7.32 \\

FlowAR-H\methodcite{flowar}
& $50{\times}2^{\dagger}$
& latent
& 1.9B
& 1.68
& 6.13
& 6.16 \\

MAR-H\methodcite{mar}
& $256{\times}2{\times}100$
& latent
& 942M
& 1.56
& 5.61
& 6.31 \\

MAR-L, DeTok\methodcite{detok}
& $256{\times}2{\times}100$
& latent
& 478M
& \textbf{1.39}
& \textbf{5.49}
& \textbf{6.05} \\

\multicolumn{7}{@{}l}{
  \quad\textit{with semantic distillation}
} \\

REG\methodcite{reg}
& $250{\times}2^{\dagger}$
& latent
& 685M
& 1.54
& 4.64
& 5.15 \\

SiT-XL/2-REPA\methodcite{repa}
& $250{\times}2^{\dagger}$
& latent
& 675M
& 1.42
& 5.45
& 6.05 \\

LightningDiT\methodcite{lightningdit}
& $250{\times}2$
& latent
& 675M
& 1.42
& 4.57
& 5.02 \\

DDT-XL\methodcite{ddt}
& $250{\times}2$
& latent
& 675M
& 1.26
& 5.70
& 6.38 \\

REPA-E\methodcite{repae}
& $250{\times}2^{\dagger}$
& latent
& 676M
& 1.17
& \textbf{3.04}
& \textbf{3.33} \\

RAE-XL\methodcite{rae}
& $50{\times}2^{\dagger}$
& latent
& 839M
& \textbf{1.16}
& 3.26
& 3.92 \\

\midrule

\multicolumn{7}{@{}l}{
  \textit{latent-space models, one-step}
} \\

Drift-L (latent)\methodcite{drift}
& 1
& latent
& 463M
& 1.53
& 10.92
& 11.32 \\

iMF-XL\methodcite{imf}
& 1
& latent
& 610M
& 1.82
& 8.39
& 8.72 \\

iMF-XL\methodcite{imf}
& 2
& latent
& 610M
& 1.61
& 7.48
& \textbf{7.79} \\


\midrule

\multicolumn{7}{@{}l}{
  \textit{pixel-space models, multi-step backbones}
} \\


JiT-B\methodcite{li2026jit}
& $50{\times}2{\times}2^{\dagger}$
& pixel
& 131M
& 3.71
& 15.65
& 15.06 \\

\rowcolor{fdrow}
\quad $+$ FD-loss\methodcite{yang2026representation}
& 1
& pixel
& 131M
& 1.00
& 5.53
& 8.45 \\

\rowcolor{oursrow}
\quad $+$ AdvFD (Ours)
& 1
& pixel
& 131M
& \textbf{0.79}
& \textbf{3.92}
& \textbf{6.03} \\

JiT-L\methodcite{li2026jit}
& $50{\times}2{\times}2^{\dagger}$
& pixel
& 459M
& 2.59
& 10.73
& 10.27 \\

\rowcolor{fdrow}
\quad $+$ FD-loss\methodcite{yang2026representation}
& 1
& pixel
& 459M
& 0.77
& 3.24
& 5.46 \\

\rowcolor{oursrow}
\quad $+$ AdvFD (Ours)
& 1
& pixel
& 459M
& \textbf{0.73}
& \textbf{2.01}
& \textbf{3.20} \\

JiT-H\methodcite{li2026jit}
& $50{\times}2{\times}2^{\dagger}$
& pixel
& 953M
& 1.97
& 7.66
& 9.07 \\

\rowcolor{fdrow}
\quad $+$ FD-loss\methodcite{yang2026representation}
& 1
& pixel
& 953M
& 0.75
& 2.65
& 4.44 \\

\rowcolor{oursrow}
\quad $+$ AdvFD (Ours)
& 1
& pixel
& 953M
& \textbf{0.72}
& \textbf{1.80}
& \textbf{2.93} \\

\midrule

\multicolumn{7}{@{}l}{
  \textit{pixel-space models, one-step}
} \\

Drift-L (pixel)\methodcite{drift}
& 1
& pixel
& 465M
& 1.43
& 10.51
& 11.18 \\

pMF-B\methodcite{lu2026meanflow}
& 1
& pixel
& 118M
& 3.31
& 13.70
& 11.82 \\

\rowcolor{fdrow}
\quad $+$ FD-loss\methodcite{yang2026representation}
& 1
& pixel
& 118M
& 0.85
& 3.50
& 4.49 \\

\rowcolor{oursrow}
\quad $+$ AdvFD (Ours)
& 1
& pixel
& 118M
& \textbf{0.81}
& \textbf{3.32}
& \textbf{4.22} \\

pMF-L\methodcite{lu2026meanflow}
& 1
& pixel
& 410M
& 2.72
& 9.09
& 7.62 \\

\rowcolor{fdrow}
\quad $+$ FD-loss\methodcite{yang2026representation}
& 1
& pixel
& 410M
& 0.78
& 2.09
& 2.72 \\

\rowcolor{oursrow}
\quad $+$ AdvFD (Ours)
& 1
& pixel
& 410M
& \textbf{0.77}
& \textbf{1.89}
& \textbf{2.57} \\

pMF-H\methodcite{lu2026meanflow}
& 1
& pixel
& 935M
& 2.29
& 6.87
& 6.09 \\

\rowcolor{fdrow}
\quad $+$ FD-loss\methodcite{yang2026representation}
& 1
& pixel
& 935M
& 0.77
& 1.89
& 2.69 \\

\rowcolor{oursrow}
\quad $+$ AdvFD (Ours)
& 1
& pixel
& 935M
& \textbf{0.74}
& \textbf{1.74}
& \textbf{2.50} \\

\bottomrule
\end{tabular}%
}

\end{table*}

\subsection{Ablation Study}

\begin{table}[t]
\centering
\small
\caption{
\textbf{Alternative anti-hacking strategies and whitening ablation.}
All experiments use JiT-B with the SIM FD objective. Lower is better for all metrics.
}

\begin{minipage}[t]{0.48\linewidth}
\centering
\small
\setlength{\tabcolsep}{5pt}
\renewcommand{\arraystretch}{1.08}
\begin{tabular}{@{}lccc@{}}
\toprule
Method
& FID $\downarrow$
& FD-r6 $\downarrow$
& FD-r3 $\downarrow$
\tabularnewline
\midrule
FD-loss
& 1.00
& 5.53
& 8.45
\tabularnewline
FD-loss + PatchGAN
& 0.90
& 5.57
& 8.90
\tabularnewline
FD-loss + DMD
& 1.77
& 5.90
& 6.75
\tabularnewline
\midrule
AdvFD
& \textbf{0.79}
& \textbf{3.92}
& \textbf{6.03}
\tabularnewline
\bottomrule
\end{tabular}

\vspace{4pt}
\raggedright
\textbf{(a) Alternative remedies.}
External guidance yields inconsistent gains, whereas AdvFD improves all metrics.
\end{minipage}
\hfill
\begin{minipage}[t]{0.48\linewidth}
\centering
\small
\setlength{\tabcolsep}{5pt}
\renewcommand{\arraystretch}{1.08}
\begin{tabular}{@{}llccc@{}}
\toprule
Method
& Branch
& FID $\downarrow$
& FD-r6 $\downarrow$
& FD-r3 $\downarrow$
\tabularnewline
\midrule
FD-loss
& --
& 1.00
& 5.53
& 8.45
\tabularnewline
\midrule
AdvFD
& None
& 10.69
& 58.50
& 40.54
\tabularnewline
& Static
& 13.49
& 28.32
& 21.66
\tabularnewline
& Adaptive
& \textbf{0.79}
& \textbf{3.92}
& \textbf{6.03}
\tabularnewline
\bottomrule
\end{tabular}

\vspace{4pt}
\raggedright
\textbf{(b) Whitening location.}
Whitening the adversarial representation is necessary for stable optimization.
\end{minipage}

\label{tab:anti-hacking-whitening}
\end{table}

\begin{figure}[t]
\centering
\includegraphics[width=0.9\linewidth]{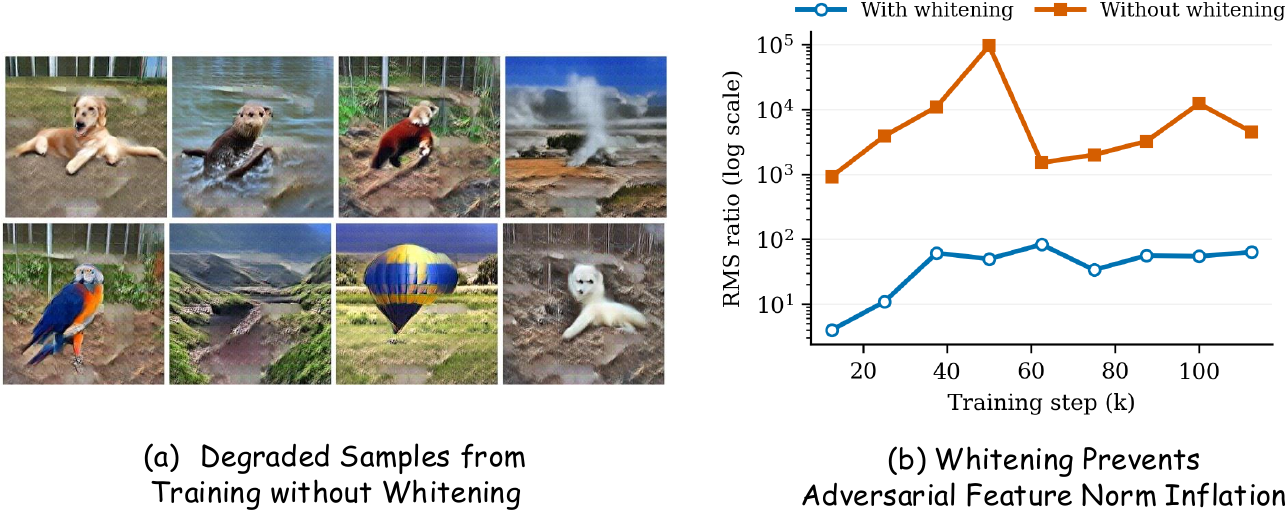}
\caption{\textbf{Effect of feature whitening on adversarial feature stability.} (a) Omitting whitening leads to severe degradation in generated samples. (b) The ratio between the pre-whitening RMS norms of the adaptive and frozen-reference features over training, shown on a logarithmic scale. Real-feature whitening removes the direct incentive to increase the adaptive Fréchet discrepancy through global feature scaling, thereby substantially suppressing feature-norm inflation in practice.}
\label{fig:whiten}
\end{figure}

\textbf{Ablation on Anti-Hacking Strategies.} We compare AdvFD with several alternative remedies for mitigating Fréchet hacking, including external PatchGAN~\citep{isola2017image} and DMD-style~\citep{yin2024one} guidance. Motivated by ASD~\citep{wei2024adversarial} and VSD~\citep{wang2023prolificdreamer}, DMD actually adopts a GAN-like adversarial learning scheme, where a learnable critic is introduced to provide adaptive feedback against artifacts overlooked by the fixed distillation objective. These methods introduce additional realism signals outside the original FD objective, but they do not necessarily improve Fréchet alignment when evaluated with feature encoders not used for training. As shown in Table~\ref{tab:anti-hacking-whitening} (a), external PatchGAN even degrades FD-r6 and FD-r3 with the FD-loss baseline, while DMD hurts FID and brings limited gains on held-out FD-r3. In contrast, AdvFD with an Inception-initialized adversarial representation substantially improves FD-r6 and FD-r3, suggesting that calibrated adversarial representations better expose blind spots of static FD objectives than external score-based remedies.

\textbf{Ablation on Whitening Location.}
Table~\ref{tab:anti-hacking-whitening} (b) shows that whitening must be applied to the adversarial representation. Without whitening, adversarial optimization becomes unstable, and FID, FD-r6, and FD-r3 deteriorate to 10.69, 58.50, and 40.54, respectively. Whitening only the static representations still leaves the trainable branch unconstrained and therefore provides limited stabilization. In contrast, whitening the adversarial representation directly suppresses feature-scale inflation, stabilizes training, and improves all metrics over the FD-Loss baseline. This result is further supported by Figure~\ref{fig:whiten}: panel (a) shows the severe degradation in generated samples without whitening, while panel (b) shows that the pre-whitening RMS norm of the adaptive features grows by several orders of magnitude relative to the frozen reference. With whitening, this ratio remains substantially lower and more stable, confirming that whitening prevents the adaptive branch from increasing the Fréchet discrepancy mainly through global feature scaling.

\textbf{Ablation on adversarial representation backbones and initialization.}
Table~\ref{tab:adv-arch} varies the backbone and initialization of $\psi_\omega$ while keeping the static SIM-FD objective unchanged. SigLIP and MAE are adapted with rank-16 LoRA~\citep{hu2022lora} due to memory constraints, whereas Inception is fully fine-tuned.

Pretrained Inception performs best across all metrics. Pretrained initialization provides a well-conditioned feature space from the start, letting the $D$-step focus on residual real--generated discrepancies, whereas a randomly initialized encoder must first learn basic visual features and more readily maximizes FD through low-level shortcuts. SigLIP and MAE underperform because patchification and global aggregation reduce sensitivity to the fine-grained textures and local artifacts that characterize generation failures. Initializing from Inception may appear to contradict Section~\ref{sec:motivation}, which identifies that space as the most readily exploited, but the two settings differ in what is held fixed: a frozen encoder exposes a permanent blind subspace, whereas the $D$-step re-selects the comparison geometry as $q_\theta$ evolves, so an exploitable direction is re-weighted rather than left unpenalized. What our diagnostic identifies is staticity, not the architecture. Table~\ref{tab:wall-clock} confirms this: freezing the same additional Inception branch degrades all metrics, while making it trainable improves all three.

\begin{table}[t]
\centering
\caption{
\textbf{Ablation on adversarial representation backbones and initialization.}
We keep the static training objective fixed to SIM FD and vary the backbone and initialization of the adversarial representation $\psi_{\omega}$. ``Pretrained'' denotes initialization from the corresponding pretrained visual encoder, while ``Random'' uses the same architecture without loading pretrained weights. FD-r3 excludes the SIM representations used for training and evaluates generalization to held-out feature spaces. Lower is better for all metrics. Best results are shown in \textbf{bold}, and second-best results are \underline{underlined}.
}
\label{tab:adv-arch}
\resizebox{0.85\linewidth}{!}{
\begin{tabular}{lllccc}
\toprule
Method
& Adv. Repr.
& Initialization
& FID $\downarrow$
& FD-r6 $\downarrow$
& FD-r3 $\downarrow$ \\
\midrule
FD-loss (Baseline)
& --
& --
& 1.00
& 5.53
& 8.45 \\
AdvFD
& SigLIP
& Pretrained
& 0.89
& \underline{4.55}
& \underline{7.40} \\
AdvFD
& MAE
& Pretrained
& 0.90
& 4.73
& 7.42 \\
AdvFD
& Inception
& Random
& \underline{0.85}
& 5.69
& 7.63 \\
AdvFD
& Inception
& Pretrained
& \textbf{0.79}
& \textbf{3.92}
& \textbf{6.03} \\
\bottomrule
\end{tabular}
}
\end{table}

\section{Conclusion}
\label{sec:conclusion}

We presented \textbf{AdvFD}, a calibrated adversarial representation training framework that improves the robustness of Fréchet-based distribution matching under static representation targets. By adaptively exposing residual discrepancies overlooked by static feature spaces, AdvFD improves one-step ImageNet generation across different backbones while preserving the same 1-NFE sampling cost. AdvFD consistently enhances both the optimized Fréchet metrics and those measured in feature spaces not used during training, demonstrating improved robustness beyond the target representation. We hope this work encourages further exploration of adaptive comparison spaces for reliable distribution-level training and its extension to larger-scale, multimodal, and video generation.

\bibliography{iclr2026_conference}
\bibliographystyle{iclr2026_conference}

\clearpage
\newpage
\appendix

\section{Finite-Horizon Boundedness of Representation Adaptation}
\label{app:finite-step-boundedness}

We briefly show that the finite, gradient-clipped AdamW~\citep{loshchilov2017decoupled} updates used in AdvFD remain within a bounded neighborhood of their initialization. Let $\tilde{g}^{(k)}=\operatorname{clip}_{\tau}(g^{(k)})$, such that $\|\tilde{g}^{(k)}\|_2\leq\tau$. Starting from zero-initialized moments, the bias-corrected first moment of AdamW is a convex combination of the clipped gradients and therefore satisfies $\|\hat{m}^{(k)}\|_2\leq\tau$. Since $\hat{v}^{(k)}$ is elementwise nonnegative and $\varepsilon_{\mathrm{A}}>0$, the preconditioned update direction is bounded as $\|u^{(k)}\|_2\leq B_{\mathrm{A}}\coloneqq\tau/\varepsilon_{\mathrm{A}}$.

For the AdamW update $\omega^{(k+1)}=(1-\eta_{\mathrm{D}}\lambda_{\mathrm{wd}})\omega^{(k)}+\eta_{\mathrm{D}}u^{(k+1)}$, where $0\leq\eta_{\mathrm{D}}\lambda_{\mathrm{wd}}\leq1$, unrolling the recursion gives, for every $k\leq K$,
\begin{equation}
\left\|\omega^{(k)}-\omega^{(0)}\right\|_2\leq\rho_K^{\mathrm{AdamW}}\coloneqq
\begin{cases}
\left[1-\left(1-\eta_{\mathrm{D}}\lambda_{\mathrm{wd}}\right)^K\right]\left(\left\|\omega^{(0)}\right\|_2+\dfrac{B_{\mathrm{A}}}{\lambda_{\mathrm{wd}}}\right), & \lambda_{\mathrm{wd}}>0,\\[6pt]
K\eta_{\mathrm{D}}B_{\mathrm{A}}, & \lambda_{\mathrm{wd}}=0.
\end{cases}
\end{equation}
Thus, the first $K$ iterates lie in a compact reachable set. Assuming compact input support and continuity of the representation network, the regularized adaptive Fréchet discrepancy is continuous on this set and therefore attains a finite maximum.

This result establishes only the finite-horizon boundedness of the implemented optimizer trajectory. It does not eliminate the scale degeneracy of raw Fréchet distance, which is addressed by real-feature whitening.

\section{Derivation and Affine Invariance of Real-Feature Whitening}
\label{app:whitening-derivation}

This appendix provides the detailed justification for the real-feature whitening used in Section~\ref{sec:AdvFD}. We first explain why the adaptive discrepancy should be invariant to invertible affine reparameterizations, then derive real-feature whitening as a solution to this requirement, and finally quantify the approximation error introduced by covariance regularization.

\paragraph{Why affine invariance is required.}
Let $\psi_{\omega}(x)\in\mathbb{R}^{d}$ denote the row-vector output of the adversarial representation. An invertible affine reparameterization
\begin{equation}
\label{eq:app-affine-equivalence}
\psi_{\omega}'(x)=\psi_{\omega}(x)A+b,
\qquad
A\in\mathrm{GL}(d),
\qquad
b\in\mathbb{R}^{d},
\end{equation}
changes the origin, scale, and axes of the feature coordinates without changing the information represented by $\psi_{\omega}$. An adaptive discrepancy should therefore assign the same value to affine-equivalent representations:
\begin{equation}
\label{eq:app-affine-invariance-requirement}
\mathcal{D}\left(p,q_{\theta};\psi_{\omega}A+b\right)=\mathcal{D}\left(p,q_{\theta};\psi_{\omega}\right).
\end{equation}

Raw Fr\'echet distance does not satisfy this requirement because it depends on the Euclidean coordinates through the feature mean and covariance. In particular, feature scaling is obtained by setting $A=cI$ and $b=0$. For any $c>0$, the corresponding feature statistics satisfy
\begin{equation}
\mu_{r}^{c\psi}=c\mu_{r}^{\psi},
\qquad
\Sigma_{r}^{c\psi}=c^{2}\Sigma_{r}^{\psi},
\qquad
r\in\{p,q_{\theta}\},
\end{equation}
and hence
\begin{equation}
\label{eq:app-scale-degeneracy}
D_{\mathrm{FD}}^{c\psi_{\omega}}\left(p,q_{\theta}\right)=c^{2}D_{\mathrm{FD}}^{\psi_{\omega}}\left(p,q_{\theta}\right).
\end{equation}
Whenever the original discrepancy is nonzero,
\begin{equation}
\label{eq:app-unbounded-raw-fd}
\sup_{c>0}D_{\mathrm{FD}}^{c\psi_{\omega}}\left(p,q_{\theta}\right)=+\infty.
\end{equation}
Thus, increasing raw FD can be achieved by continually expanding the feature coordinates, without exposing any additional difference between the represented distributions. Equation~\ref{eq:app-affine-invariance-requirement} removes this coordinate-dependent direction from the adaptive objective.

\paragraph{Real-feature whitening as a solution.}
Let
\begin{equation}
\label{eq:app-real-statistics}
\mu_{p}^{\psi}=\mathbb{E}_{x\sim p}\left[\psi_{\omega}(x)\right],
\qquad
\Sigma_{p}^{\psi}=\operatorname{Cov}_{x\sim p}\left[\psi_{\omega}(x)\right]
\end{equation}
denote the population mean and covariance of the real features. We consider a standardized representation of the form
\begin{equation}
\label{eq:app-standardized-representation}
\bar{\psi}_{\omega}(x)=\left(\psi_{\omega}(x)-\mu_{p}^{\psi}\right)W_{\omega}.
\end{equation}
Centering removes the dependence on the coordinate origin. To remove the scale and covariance geometry, we require the transformed real covariance to be identity:
\begin{equation}
\label{eq:app-whitening-condition}
W_{\omega}^{\top}\Sigma_{p}^{\psi}W_{\omega}=I.
\end{equation}
When $\Sigma_{p}^{\psi}\succ0$, every solution of Equation~\ref{eq:app-whitening-condition} can be written as
\begin{equation}
W_{\omega}=\left(\Sigma_{p}^{\psi}\right)^{-1/2}R,
\qquad
R^{\top}R=I.
\end{equation}
The remaining matrix $R$ is an arbitrary orthogonal rotation. Because FD is invariant to a common orthogonal transformation, all such solutions induce the same discrepancy. We therefore select the symmetric whitener $R=I$:
\begin{equation}
\label{eq:app-exact-real-whitening}
\bar{\psi}_{\omega}(x)=\left(\psi_{\omega}(x)-\mu_{p}^{\psi}\right)\left(\Sigma_{p}^{\psi}\right)^{-1/2}.
\end{equation}
The resulting real-feature statistics are
\begin{equation}
\mu_{p}^{\bar{\psi}}=0,
\qquad
\Sigma_{p}^{\bar{\psi}}=I.
\end{equation}

It remains to verify that this construction satisfies Equation~\ref{eq:app-affine-invariance-requirement}. Under the affine transformation in Equation~\ref{eq:app-affine-equivalence}, the real-feature statistics become
\begin{equation}
\mu_{p}'=\mu_{p}^{\psi}A+b,
\qquad
\Sigma_{p}'=A^{\top}\Sigma_{p}^{\psi}A.
\end{equation}
Whitening the transformed representation using its own real statistics gives
\begin{align}
\bar{\psi}_{\omega}'(x)
&=\left(\psi_{\omega}'(x)-\mu_{p}'\right)\left(\Sigma_{p}'\right)^{-1/2}\\
&=\left(\psi_{\omega}(x)-\mu_{p}^{\psi}\right)A\left(A^{\top}\Sigma_{p}^{\psi}A\right)^{-1/2}\\
&=\bar{\psi}_{\omega}(x)Q_{A},
\label{eq:app-whitened-affine-equivalence}
\end{align}
where
\begin{equation}
\label{eq:app-orthogonal-factor}
Q_{A}=\left(\Sigma_{p}^{\psi}\right)^{1/2}A\left(A^{\top}\Sigma_{p}^{\psi}A\right)^{-1/2}.
\end{equation}
This matrix is orthogonal because
\begin{equation}
Q_{A}^{\top}Q_{A}
=\left(A^{\top}\Sigma_{p}^{\psi}A\right)^{-1/2}
\left(A^{\top}\Sigma_{p}^{\psi}A\right)
\left(A^{\top}\Sigma_{p}^{\psi}A\right)^{-1/2}
=I.
\end{equation}
Thus, affine-equivalent raw representations become orthogonally equivalent after real-feature whitening. Since FD is invariant to a common orthogonal transformation,
\begin{equation}
\label{eq:app-exact-affine-invariance}
D_{\mathrm{FD}}^{\bar{\psi}_{\omega}'}\left(p,q_{\theta}\right)=D_{\mathrm{FD}}^{\bar{\psi}_{\omega}}\left(p,q_{\theta}\right),
\end{equation}
which proves Equation~\ref{eq:app-affine-invariance-requirement}. For the scaling transformation $A=cI$ with $c>0$, $Q_{A}=I$, so the whitened features themselves remain unchanged. Consequently, the adversarial representation cannot increase the whitened discrepancy merely by expanding its global coordinate scale.

Under exact whitening, the adaptive discrepancy can be written as
\begin{equation}
\label{eq:app-whitened-adaptive-fd}
D_{\mathrm{adv}}\left(p,q_{\theta};\omega\right)=\left\|\mu_{q}^{\bar{\psi}}\right\|_{2}^{2}+\operatorname{Tr}\left[I+\Sigma_{q}^{\bar{\psi}}-2\left(\Sigma_{q}^{\bar{\psi}}\right)^{1/2}\right],
\end{equation}
where $\mu_{q}^{\bar{\psi}}$ and $\Sigma_{q}^{\bar{\psi}}$ are the mean and covariance of the whitened generated features.

\paragraph{Approximation error introduced by covariance regularization.}
Exact whitening assumes population statistics and $\Sigma_{p}^{\psi}\succ0$. In practice, the real-feature statistics are estimated from a minibatch, and the empirical covariance may be ill-conditioned or rank deficient. We therefore use
\begin{equation}
\label{eq:app-regularized-whitening}
\bar{\psi}_{\omega,\epsilon}(x)=\left(\psi_{\omega}(x)-\hat{\mu}_{p}^{\psi}\right)\left(\hat{\Sigma}_{p}^{\psi}+\epsilon I\right)^{-1/2},
\qquad
\epsilon>0.
\end{equation}
The ridge makes the inverse covariance square root well defined and bounds its operator norm:
\begin{equation}
\left\|\left(\hat{\Sigma}_{p}^{\psi}+\epsilon I\right)^{-1/2}\right\|_{2}\leq\epsilon^{-1/2}.
\end{equation}

To quantify the deviation from exact whitening, let
\begin{equation}
\hat{\Sigma}_{p}^{\psi}=U\operatorname{diag}\left(\lambda_{1},\ldots,\lambda_{d}\right)U^{\top}.
\end{equation}
The covariance of the regularized-whitened real features is
\begin{align}
\Sigma_{p,\epsilon}^{\bar{\psi}}
&=\left(\hat{\Sigma}_{p}^{\psi}+\epsilon I\right)^{-1/2}\hat{\Sigma}_{p}^{\psi}\left(\hat{\Sigma}_{p}^{\psi}+\epsilon I\right)^{-1/2}\\
&=U\operatorname{diag}\left(\frac{\lambda_{1}}{\lambda_{1}+\epsilon},\ldots,\frac{\lambda_{d}}{\lambda_{d}+\epsilon}\right)U^{\top}.
\label{eq:app-regularized-real-covariance}
\end{align}
Exact whitening would map every nonzero covariance eigenvalue to one. With regularization, the error along the $i$-th eigenvector is
\begin{equation}
\label{eq:app-directional-whitening-error}
\left|1-\frac{\lambda_{i}}{\lambda_{i}+\epsilon}\right|=\frac{\epsilon}{\lambda_{i}+\epsilon}.
\end{equation}
Therefore, on any covariance subspace satisfying $\lambda_i\geq\lambda_{\min}>0$,
\begin{equation}
\label{eq:app-whitening-error-bound}
\left\|\Sigma_{p,\epsilon}^{\bar{\psi}}-I\right\|_{2}
\leq\frac{\epsilon}{\lambda_{\min}+\epsilon}
\leq\frac{\epsilon}{\lambda_{\min}}.
\end{equation}
The relative change in the whitening multiplier is similarly bounded by
\begin{equation}
\label{eq:app-feature-scaling-error}
1-\frac{(\lambda_i+\epsilon)^{-1/2}}{\lambda_i^{-1/2}}
=1-\sqrt{\frac{\lambda_i}{\lambda_i+\epsilon}}
\leq\frac{\epsilon}{2\lambda_i}.
\end{equation}
Thus, the regularized transformation differs from exact whitening by $\mathcal{O}(\epsilon/\lambda_{\min})$ on well-conditioned covariance directions. Directions with very small empirical eigenvalues incur a larger approximation error but are intentionally regularized to prevent unstable inverse scaling.

We use $\epsilon=10^{-3}$ in all experiments. Equation~\ref{eq:app-directional-whitening-error} gives a covariance error below $1\%$ for directions with $\lambda_i\geq0.1$ and below $0.1\%$ for directions with $\lambda_i\geq1$. Because $\epsilon>0$ breaks exact affine invariance, the implemented transformation should be viewed as a regularized approximation to the canonicalization derived above. Exact affine invariance is recovered in the full-rank limit as $\epsilon\rightarrow0$.

\paragraph{Deviation from detached EMA statistics.}
The cancellation above further assumes that whitening uses the real-feature statistics of the current $\psi_\omega$. We estimate them with an EMA that is detached from the $D$-step gradient, so an instantaneous rescaling of $\psi_\omega$ is not cancelled within the same step and is corrected only once the EMA catches up. The resulting deviation is governed by the drift of $\Sigma^{\psi}_{p}$ over that timescale, which is small in our setting because $\eta_D \le 2\times 10^{-6}$ and $\psi_\omega$ is updated only every two generator steps. Together with the $O(\epsilon/\lambda_{\min})$ term above, invariance therefore holds up to a controlled approximation rather than exactly, which is consistent with the bounded residual growth in Figure~\ref{fig:whiten} (b).

\section{Transport-Based Relation to GANs}
\label{app:relation-to-gans}

We provide a transport-based interpretation of the relation between conventional adversarial training, WGAN, and AdvFD. The central distinction is whether the inner optimization learns a function under a predefined distribution geometry or learns the geometry in which the distributions are compared.

\paragraph{From GAN discrimination to distribution transport.}
The original GAN~\citep{goodfellow2020generative} optimizes
\begin{equation}
\label{eq:app-gan-objective}
\min_{\theta}
\sup_{D}
\left[
\mathbb{E}_{x\sim p}
\log D(x)
+
\mathbb{E}_{x\sim q_{\theta}}
\log
\left(
1-D(x)
\right)
\right].
\end{equation}
For a static generated distribution $q_{\theta}$, the optimal discriminator is
\begin{equation}
\label{eq:app-optimal-discriminator}
D^{\star}(x)
=
\frac{p(x)}
{p(x)+q_{\theta}(x)}.
\end{equation}
Substituting $D^{\star}$ into Equation~\ref{eq:app-gan-objective} gives
\begin{equation}
\label{eq:app-gan-js}
\sup_{D}
V
\left(
D;p,q_{\theta}
\right)
=
-\log 4
+
2D_{\mathrm{JS}}
\left(
p\|q_{\theta}
\right).
\end{equation}
Thus, the discriminator can be analytically eliminated, and the resulting objective measures distributional overlap through the Jensen--Shannon divergence. This formulation does not introduce a cost describing how probability mass should move between distinct sample locations.

\paragraph{WGAN under a fixed transport geometry.}
WGAN~\citep{arjovsky2017wasserstein} instead begins by fixing a ground cost $d_{\mathcal{X}}(x,y)$ between samples in the input space. This ground cost defines the Wasserstein-1 transport problem
\begin{equation}
\label{eq:app-wgan-primal}
W_{1}
\left(
p,q_{\theta}
\right)
=
\inf_{\gamma\in\Pi(p,q_{\theta})}
\mathbb{E}_{(x,y)\sim\gamma}
\left[
d_{\mathcal{X}}(x,y)
\right],
\end{equation}
where $\Pi(p,q_{\theta})$ denotes the set of couplings whose marginals are $p$ and $q_{\theta}$. The coupling $\gamma$ specifies how probability mass is transported, while $d_{\mathcal{X}}$ specifies the cost of moving it.

Once $d_{\mathcal{X}}$ is fixed, the resulting optimal-transport problem can be expressed through the Kantorovich--Rubinstein dual:
\begin{equation}
\label{eq:app-wgan-dual}
W_{1}
\left(
p,q_{\theta}
\right)
=
\sup_{\|f\|_{\mathrm{Lip}}\leq 1}
\left[
\mathbb{E}_{x\sim p}f(x)
-
\mathbb{E}_{x\sim q_{\theta}}f(x)
\right].
\end{equation}
The WGAN critic $f$ is therefore a scalar dual potential for a transport problem whose sample representation and ground cost have already been specified. The critic optimization solves the dual problem but does not change how distances between samples are defined.

In summary, WGAN follows the structure
\begin{equation}
\label{eq:app-wgan-structure}
\underbrace{
d_{\mathcal{X}}(x,y)
}_{\text{fixed transport geometry}}
\quad\Longrightarrow\quad
\underbrace{
\sup_{\|f\|_{\mathrm{Lip}}\leq 1}
\left[
\mathbb{E}_{p}f-\mathbb{E}_{q_{\theta}}f
\right]
}_{\text{learned dual transport potential}}.
\end{equation}

\paragraph{Fréchet distance as Gaussian transport.}
AdvFD begins from a different transport functional. For a representation
$\psi_{\omega}:\mathcal{X}\rightarrow\mathbb{R}^{d}$, define the corresponding pushforward feature distributions as
\begin{equation}
\label{eq:app-feature-pushforwards}
P_{\omega}
=
(\psi_{\omega})_{\#}p,
\qquad
Q_{\theta,\omega}
=
(\psi_{\omega})_{\#}q_{\theta}.
\end{equation}
Let
\begin{equation}
\label{eq:app-gaussian-moment-map}
\mathcal{G}_{2}(\nu)
=
\mathcal{N}
\left(
\mu_{\nu},
\Sigma_{\nu}
\right)
\end{equation}
denote the Gaussian distribution with the same mean and covariance as $\nu$. For a static representation, the Fréchet discrepancy is the squared Wasserstein-2 distance between the two Gaussian moment models:
\begin{equation}
\label{eq:app-fd-as-gaussian-w2}
\begin{aligned}
D_{\mathrm{FD}}^{\psi_{\omega}}
\left(
p,q_{\theta}
\right)
&=
W_{2}^{2}
\left(
\mathcal{G}_{2}(P_{\omega}),
\mathcal{G}_{2}(Q_{\theta,\omega})
\right) \\
&=
\left\|
\mu_{p}^{\psi}
-
\mu_{q}^{\psi}
\right\|_{2}^{2} \\
&\quad+
\operatorname{Tr}
\left[
\Sigma_{p}^{\psi}
+
\Sigma_{q}^{\psi}
-
2
\left(
(\Sigma_{p}^{\psi})^{1/2}
\Sigma_{q}^{\psi}
(\Sigma_{p}^{\psi})^{1/2}
\right)^{1/2}
\right].
\end{aligned}
\end{equation}
Equivalently,
\begin{equation}
\label{eq:app-gaussian-transport-primal}
D_{\mathrm{FD}}^{\psi_{\omega}}
\left(
p,q_{\theta}
\right)
=
\inf_{\gamma\in
\Pi\left(
\mathcal{G}_{2}(P_{\omega}),
\mathcal{G}_{2}(Q_{\theta,\omega})
\right)}
\mathbb{E}_{(z,z')\sim\gamma}
\left[
\|z-z'\|_{2}^{2}
\right].
\end{equation}
Unlike WGAN, this Gaussian transport problem has a closed-form optimal value once $\psi_{\omega}$ is fixed. The mean term in Equation~\ref{eq:app-fd-as-gaussian-w2} measures the cost of translating the center of the generated feature distribution, while the covariance term measures the cost of aligning its scale and orientation with those of the real feature distribution.

\paragraph{AdvFD as learned transport geometry.}
AdvFD does not learn a dual potential for one fixed transport problem. Instead, it optimizes the representation that determines the feature distributions participating in the Gaussian transport problem. Let $\Psi$ denote the set of representations reachable by the $D$-step from the initialization $\psi_{\omega_0}$ under the clipping constraint of \ref{eq:d-step-gradient}. We have the optimization goal:
\begin{equation}
\sup_{\psi_\omega \in \Psi} D^{\psi_\omega}_{\mathrm{FD}}(p,q_\theta) = \sup_{\psi_\omega \in \Psi} W_2^2\big(\mathcal{G}_2(P_\omega),\, \mathcal{G}_2(Q_{\theta,\omega})\big).
\end{equation}
Each representation induces a feature-space comparison cost
\begin{equation}
\label{eq:app-induced-feature-cost}
d_{\psi_{\omega}}(x,y)
=
\left\|
\psi_{\omega}(x)
-
\psi_{\omega}(y)
\right\|_{2},
\end{equation}
and changes the mean and covariance structures used in Equation~\ref{eq:app-fd-as-gaussian-w2}. Maximizing over $\psi_{\omega}$ therefore searches for a feature geometry in which the remaining second-order mismatch between $p$ and $q_{\theta}$ is most pronounced.

For each fixed $\psi_{\omega}$, the Gaussian transport problem is solved in closed form. The remaining representation optimization generally has no analytic solution and is performed through the D-step. The generator then minimizes the resulting discrepancy through the G-step. Hence, AdvFD follows the structure
\begin{equation}
\label{eq:app-adafd-structure}
\underbrace{\bar\psi_{\omega_t}\ \longrightarrow\ \bar\psi_{\omega_{t+1}}}_{\text{adversely updated feature geometry}}\ \ \text{applied to}\ \ \underbrace{W_2^2\big(\mathcal G_2(P_\omega),\mathcal G_2(Q_{\theta,\omega})\big)}_{\text{fixed closed-form Gaussian transport functional}}
\end{equation}

The contrast with WGAN can therefore be summarized as
\begin{equation}
\label{eq:app-wgan-adafd-summary}
\begin{aligned}
\text{WGAN:}\qquad
&
\text{fixed transport geometry}
+
\text{learned dual transport potential}, \\
\text{AdvFD:}\qquad
&
\text{learned feature geometry}
+
\text{fixed Gaussian transport functional}.
\end{aligned}
\end{equation}
WGAN learns the solution to an optimal-transport problem under a predefined sample-space geometry. AdvFD instead learns the geometry in which a predefined second-order transport objective is evaluated.

\section{More Experiment Results}

\textbf{Ablation on Adversarial Loss Weight.} Table~\ref{tab:adv-weight} studies the effect of the adaptive loss weight $\lambda_{\mathrm{adv}}$. All nonzero weights improve over the FD-loss baseline, confirming that the adversarial representation provides complementary supervision. However, a small weight such as $0.01$ underutilizes the exposed discrepancies and brings relatively limited gains, particularly on the FD-r3 metric. Increasing $\lambda_{\mathrm{adv}}$ to $0.05$ or $0.10$ yields substantially better results, with $0.10$ achieving the best FD-r6 and FD-r3. A larger weight of $0.20$ degrades all metrics, as the adaptive term begins to dominate the stable static FD objective and may introduce overly strong or conflicting gradients that disrupt the original post-training process. These results suggest that a moderate adaptive loss weight provides the best balance between preserving static distribution alignment and correcting its representation blind spots.

\begin{table}[t]
\centering
\caption{
\textbf{Ablation on adversarial loss weight.}
We vary the adversarial loss weight $\lambda_{\mathrm{adv}}$ under the same JiT-B post-training setting. The FD-loss baseline and all AdvFD variants are trained with the SIM FD objective. FD-r3 excludes the SIM representations used for training. Lower is better for all metrics.
}
\label{tab:adv-weight}
\resizebox{0.65\linewidth}{!}{
\begin{tabular}{lcccc}
\toprule
Method
& $\lambda_{\mathrm{adv}}$
& FID $\downarrow$
& FD-r6 $\downarrow$
& FD-r3 $\downarrow$
\tabularnewline
\midrule
FD-loss baseline
& --
& 1.00
& 5.53
& 8.45
\tabularnewline
\midrule
AdvFD
& 0.01
& 0.83
& 4.51
& 7.34
\tabularnewline
AdvFD
& 0.05
& 0.80
& 4.68
& 6.28
\tabularnewline
AdvFD
& 0.10
& \textbf{0.79}
& \textbf{3.92}
& \textbf{6.03}
\tabularnewline
AdvFD
& 0.20
& 0.87
& 5.46
& 6.74
\tabularnewline
\bottomrule
\end{tabular}
}
\end{table}

\textbf{Wall-clock-aligned comparison.}
Table~\ref{tab:wall-clock} shows that the gains of AdvFD cannot be attributed to additional computation or merely adding another representation branch. When the adversarial representation is frozen, it becomes an additional static target that cannot adjust to the evolving generated distribution, thereby overemphasizing a fixed feature geometry and substantially degrading all metrics. Likewise, extending FD-loss to match the wall-clock cost of AdvFD provides no improvement, indicating that longer optimization of the same static objectives is insufficient. In contrast, AdvFD achieves an FID of 0.79, FD-r6 of 3.92, and FD-r3 of 6.03 under the same wall-clock budget, confirming that the gains arise from adaptively updating the representation to expose residual real--generated discrepancies.

\begin{table}[t]
\centering
\caption{
\textbf{Wall-clock-aligned comparison.}
AdvFD (Frozen) retains the additional representation branch but freezes its parameters, isolating the effect of adversarial representation learning. FD-loss is additionally trained for $N$ steps to match the total wall-clock cost of AdvFD.
}
\setlength{\tabcolsep}{6.5pt}
\renewcommand{\arraystretch}{1.08}
\begin{tabular}{@{}lcccccc@{}}
\toprule
Method
& Adaptive Repr.
& Train Steps
& Wall-Clock
& FID $\downarrow$
& FD-r6 $\downarrow$
& FD-r3 $\downarrow$
\tabularnewline
\midrule
FD-loss
& --
& $T$
& $C$
& 1.00
& 5.53
& 8.45
\tabularnewline
AdvFD (Frozen)
& Frozen
& $T$
& $C_{\mathrm{AdvFD}}$
& 1.62
& 11.02
& 8.71
\tabularnewline
FD-loss
& --
& $T+N$
& $C_{\mathrm{AdvFD}}$
& 1.71
& 7.70
& 8.63
\tabularnewline
\midrule
AdvFD
& Trainable
& $T$
& $C_{\mathrm{AdvFD}}$
& \textbf{0.79}
& \textbf{3.92}
& \textbf{6.03}
\tabularnewline
\bottomrule
\end{tabular}

\label{tab:wall-clock}
\end{table}



\section{Implementation and Evaluation Details}

\subsection{Implementation Details}

\paragraph{Training Settings.}
We conduct all experiments on ImageNet-1K at a resolution of
$256\times256$. We initialize the post-training stage from the released
base checkpoints and keep the original pMF and JiT generator architectures.
Unless otherwise stated, the main experiments use the FD-SIM objective
together with the FD-Adv objective. The training hyperparameters are
summarized in Table~\ref{tab:training_settings}.

\begin{table*}[t]
    \centering
    \small
    \caption{Training hyperparameters for the $256\times256$ experiments.}
    \label{tab:training_settings}
    \begin{tabular}{lcc}
        \toprule
        Hyperparameter & pMF & JiT \\
        \midrule

        \multicolumn{3}{l}{\textbf{Generator}} \\
        \addlinespace[1pt]
        Initialization &
        \texttt{pMF-B/L/H.pth} &
        \texttt{JiT-B/L/H.pth} \\
        Global batch size & 1024 & 1024 \\
        Per-GPU batch size &
        $1024/G$ (128 with 8 GPUs) &
        $1024/G$ (128 with 8 GPUs) \\
        Training length & $100\times1250$ steps &
        $100\times1250$ steps \\
        Generator optimizer & AdamW & AdamW \\
        Generator learning rate & $1\times10^{-6}$ &
        $1\times10^{-5}$ \\
        Adam betas & $(0.9,0.95)$ & $(0.9,0.95)$ \\
        Weight decay & 0 & 0 \\
        Learning-rate warm-up & 5 epochs & 5 epochs \\
        Learning-rate schedule & cosine & cosine \\
        Sampling steps & 1 & 1 \\
        Model-weight EMA & EDM & EDM \\

        \addlinespace[4pt]
        \multicolumn{3}{l}{\textbf{Main FD Objective}} \\
        \addlinespace[1pt]
        Main FD representation &
        SigLIP + MAE + Inception &
        SigLIP + MAE + Inception \\
        FD queue size & 50,000 & 50,000 \\
        FD queue fill batch size & 256 & 256 \\
        FD-statistics EMA $\beta$ & 0.999 & 0.999 \\
        Covariance solver & \texttt{eigvalsh} & \texttt{eigvalsh} \\

        \addlinespace[4pt]
        \multicolumn{3}{l}{\textbf{Adversarial Representation}} \\
        \addlinespace[1pt]
        FD-Adv backbone & pretrained Inception &
        pretrained Inception \\
        FD-Adv learning rate & $2\times10^{-6}$ &
        $1\times10^{-6}$ \\
        FD-Adv weight & 0.05 & 0.10 \\
        FD-Adv steps per update & 1 & 1 \\
        FD-Adv update frequency &
        every 2 generator steps &
        every 2 generator steps \\
        FD-Adv start step & 1,000 & 1,000 \\
        FD-Adv warm-up steps & 4,000 & 4,000 \\
        FD-Adv gradient clipping & 1.0 & 1.0 \\
        FD-Adv real-feature gradient & detached & detached \\
        FD-Adv-statistics EMA $\beta$ & 0.99 & 0.99 \\
        FD-Adv whitening &
        enabled, $\epsilon=10^{-3}$ &
        enabled, $\epsilon=10^{-3}$ \\

        \bottomrule
    \end{tabular}
\end{table*}

The pMF models use 2D rotary position embeddings, learned positional
embeddings, and disable the auxiliary velocity head. JiT uses 2D rotary
position embeddings, learned positional embeddings, and the legacy time
convention. Gradient checkpointing is enabled for
the large-scale experiments. The generator weights are tracked with an
EDM-style model EMA, while the final released-checkpoint evaluation uses
the online generator weights.

At each training iteration, class labels are sampled uniformly from the
1000 ImageNet classes and one-step images are generated with
\texttt{num\_sampling\_steps}=1. The FD-SIM objective matches the feature
distributions extracted by SigLIP, MAE, and InceptionV3. The corresponding
feature target resolutions are 224, 224, and 299, respectively, and all
three FD terms use unit weight. The representation models are frozen with
respect to their parameters, but gradients are propagated through the
generated images.

The FD reference statistics are precomputed from ImageNet images after
center cropping and tensor conversion. During training, the FD feature
statistics are initialized from 50,000 generated samples using a queue fill
batch size of 256. The running FD statistics are then updated with an
exponential moving average using $\beta=0.999$. This EMA is independent of
the EMA used for generator weights. The generated features are gathered
across GPUs before computing their mean and covariance, and the covariance
trace term is evaluated using the symmetric eigendecomposition
implemented by \texttt{--fd\_eigvalsh}. The normalized FD loss is defined as
\[
\mathcal{L}_{\mathrm{FD}}^{\mathrm{norm}}
=
\frac{\mathcal{L}_{\mathrm{FD}}}
{\operatorname{sg}(\mathcal{L}_{\mathrm{FD}})+0.01},
\]
where $\operatorname{sg}(\cdot)$ denotes stop-gradient.

For FD-Adv training, a trainable copy of the selected representation
backbone is updated jointly with the generator. The FD-Adv critic uses AdamW with a learning rate of $2\times10^{-6}$ for pMF and $1\times10^{-6}$ for JiT, as summarized in Table~\ref{tab:training_settings}. In the SIM+Adv recipes, the critic is updated every two generator steps. The adversarial term starts at
step 1,000 and is linearly warmed up for 4,000 steps. Real-feature
gradients are detached, and the adversarial FD uses real-reference
whitening with $\epsilon=10^{-3}$. The FD-Adv feature statistics are
maintained with a separate EMA using $\beta=0.99$. The queue-size, EMA, and
backbone ablations use the same batch-size and optimizer settings, while
the shorter ablation runs use 50 training epochs.

\subsection{Evaluation Details}

We evaluate each checkpoint using 50,000 generated images. The
released-checkpoint evaluation uses an evaluation batch size of 128 images
per GPU and evaluates the online generator weights. The model-specific
sampling configurations are:
\begin{itemize}
    \item pMF-B: CFG $=8.5$, interval $[0.1,0.7]$, and noise scale $1.0$;
    \item pMF-L: CFG $=7.0$, interval $[0.2,0.7]$, and noise scale $1.0$;
    \item pMF-H: CFG $=7.0$, interval $[0.2,0.6]$, and noise scale $2.0$;
    \item JiT-B: CFG $=3.0$ and interval $[0.1,1.0]$;
    \item JiT-L: CFG $=2.4$ and interval $[0.1,1.0]$;
    \item JiT-H: CFG $=2.2$ and interval $[0.1,1.0]$.
\end{itemize}

The default evaluation suite contains InceptionV3, ConvNeXt, DINOv2-L,
MAE-L, SigLIP-SO400M, and CLIP-L. Generated images are converted to the
$[0,1]$ range before feature extraction and resized to the
encoder-specific target resolution. For each representation space, we
compute the raw Fréchet distance between the generated-image statistics and
the corresponding ImageNet reference statistics.

We report the normalized representation Fréchet distance
\[
\mathrm{FD\text{-}r}_m =
\frac{\mathrm{FD}_m}{\mathrm{valFD}_m},
\]
where the validation-set normalizers are 1.68 for Inception, 56.87 for
ConvNeXt, 14.19 for DINOv2, 0.04 for MAE, 0.60 for SigLIP, and 5.60 for
CLIP. We report three aggregate metrics. First, FDr-6 is the arithmetic
mean across all six representation spaces:
\[
\mathrm{FD\text{-}r6}
=
\frac{1}{6}\left(
\mathrm{FDr}_{\mathrm{Inception}}+
\mathrm{FDr}_{\mathrm{ConvNeXt}}+
\mathrm{FDr}_{\mathrm{DINOv2}}+
\mathrm{FDr}_{\mathrm{MAE}}+
\mathrm{FDr}_{\mathrm{SigLIP}}+
\mathrm{FDr}_{\mathrm{CLIP}}
\right).
\]
Second, FD-r3 is computed over the DINOv2, CLIP, and ConvNeXt
representation spaces:
\[
\mathrm{FD\text{-}r3}
=
\frac{1}{3}\left(
\mathrm{FDr}_{\mathrm{DINOv2}}+
\mathrm{FDr}_{\mathrm{CLIP}}+
\mathrm{FDr}_{\mathrm{ConvNeXt}}
\right).
\]
Thus, FDr-3 excludes the Inception, SigLIP, and MAE representation spaces.

\section{Limitations}
\label{sec:limitations}

Our current evaluation focuses on class-conditional ImageNet generation at $256{\times}256$ resolution. Although AdvFD consistently improves multiple generator backbones and model scales, its effectiveness on higher-resolution, text-conditioned, and video generation remains to be validated. In addition, AdvFD introduces extra training-time computation for updating the adversarial representation and computing the whitening transformation, while leaving the inference architecture and 1-NFE sampling cost unchanged.

\end{document}